\documentclass[conference]{IEEEtran}
\IEEEoverridecommandlockouts
\usepackage{cite}
\usepackage{amsmath,amssymb,amsfonts}
\usepackage{algorithmic}
\usepackage{graphicx}
\usepackage{textcomp}
\usepackage{xcolor}
\usepackage{multicol}
\usepackage{multirow}
\usepackage{array}
\usepackage{enumerate}

\usepackage{csquotes}
\usepackage{amsthm}
\usepackage{upgreek}

\newcommand{\R}{\mathbb{R}}
\newcommand{\N}{\mathcal{N}}

\DeclareMathOperator{\sgn}{sgn}

\def\BibTeX{{\rm B\kern-.05em{\sc i\kern-.025em b}\kern-.08em
    T\kern-.1667em\lower.7ex\hbox{E}\kern-.125emX}}
\begin{document}

\title{Scalable and Robust Physics-Informed Graph Neural Networks for Water Distribution Systems\\
\thanks{We gratefully acknowledge funding from the European Research Council (ERC) under the ERC Synergy Grant Water-Futures (Grant agreement No. 951424).}
}

\author{\IEEEauthorblockN{Inaam Ashraf, Andr\'e Artelt, Barbara Hammer}
\IEEEauthorblockA{\textit{Center for Cognitive Interaction Technology} \\
\textit{Bielefeld University}\\
Bielefeld, Germany \\
\{mashraf, aartelt, bhammer\}@techfak.uni-bielefeld.de }
}

\maketitle

\begin{abstract}

Water distribution systems (WDSs) are an important part of critical infrastructure becoming increasingly significant in the face of climate change and urban population growth. We propose a robust and scalable surrogate deep learning (DL) model to enable efficient planning, expansion, and rehabilitation of WDSs. Our approach incorporates an improved graph neural network architecture, an adapted physics-informed algorithm, an innovative training scheme, and a physics-preserving data normalization method. Evaluation results on a number of WDSs demonstrate that our model outperforms the current state-of-the-art DL model. Moreover, our method allows us to scale the model to bigger and more realistic WDSs. Furthermore, our approach makes the model more robust to out-of-distribution input features (demands, pipe diameters). Hence, our proposed method constitutes a significant step towards bridging the simulation-to-real gap in the use of artificial intelligence for WDSs.  
\end{abstract}

\section{Introduction}

Reliable and robust water distribution systems (WDSs) are of utmost importance for a sustainable urban society. It is predicted that by the year 2050, more than two-thirds of the world's population will live in cities \cite{owidurbanization}. Since WDSs are responsible for providing clean water to all, their significance aligns directly with UN's sustainable developmental goal (SDG) 6 -- \enquote{Clean water and sanitation for all}. Artificial intelligence (AI) has great potential for developing surrogate models and digital twins for revamping the planning, monitoring, and rehabilitation of critical infrastructure \cite{pmlr-v176-eichenberger22a,smartcities4020029,doi:10.1061/(ASCE)PS.1949-1204.0000646}. Some work has already been done in this regard \cite{doi:10.1061/(ASCE)IS.1943-555X.0000477, valerie2024}. In this work, we reduce the gap between simulation and the real world by proposing a robust surrogate DL model that can directly be used for planning, expansion, and rehabilitation of WDSs.    

A water distribution system is a network of $N_n$ nodes connected through $N_e$ links. Consumer nodes are called junctions, large water sources like lakes and rivers are reservoirs, and artificial water sources are water tanks. The majority of links in a WDS are water pipes. At some locations, different kinds of valves and pumps are used to maintain water pressure. While planning or expanding a WDS, the structure of the network needs to be designed. For rehabilitation, the task is often to find the optimal changes required to the structure or physical components of the WDS given certain constraints in the future. In both cases, the most important task is to estimate the state of the WDS. This state refers to values of pressure head at every node and water flow through every link given the demands at every consumer node and the pressure heads at the reservoirs.  

EPANET is the state-of-the-art (SOTA) hydraulic simulator that solves a system of equations to solve the task of hydraulic state estimation in WDSs \cite{rossman2020epanet}. It is highly accurate but takes considerable time when numerous simulations are required, which is often the case in the planning, expansion, and rehabilitation of WDSs. Recently, \cite{Ashraf_Strotherm_Hermes_Hammer_2024} proposed a physics-informed DL surrogate model for the hydraulic state estimation showing promising results. However, their approach has limited generalizability to unseen input features. Moreover, since they only experimented with rather small WDSs, it is unclear how well their model scales to larger WDSs. We build on their approach and provide the following key contributions:

\begin{itemize}
    \item We propose a DL surrogate model incorporating an improved GNN architecture, an adapted physics-informed algorithm, an innovative training scheme, and a physics-preserving data normalization method.
    \item We implement two complex types of links i.e. pump and pressure-reducing valve (PRV) to allow the application of the model to realistic WDSs.
    \item We evaluate our method on a number of WDSs showing great accuracy and efficiency gains.
    \item Our model performs equally well on both small and larger realistic WDSs and is robust to both unseen in-distribution and out-of-distribution input features as demonstrated by extensive experiments.
\end{itemize}

\section{Related Work}

AI and more specifically machine learning (ML) has been used to solve certain tasks in WDS. These include demand prediction \cite{WU2023104545}, leakage detection \cite{Fan2021, valerie_icpram24} and sensor fault detection \cite{valerie_sensor}. Moreover, there exist some approaches for the tasks like optimal sensor placement \cite{Candelieri_2022, su15042981}. In the real world, pressure readings from a few sensors are available. Hence, the task of estimating the state of the WDS at every node using these sparse readings is important for leakage and sensor fault detection. This task has attracted considerable attention recently as demonstrated by \cite{su15042981, ashraf2023spatial, hajgato2021pressure, Truong_2024, xing2022stateestimation}.

Graph neural networks (GNNs) are a promising DL method to deal with network structures like WDS. There exist a lot of different GNN architectures such as spectral graph convolutional neural networks (GCNs) \cite{Bruna2014SpectralNA,kipf2017semi,defferrard2016convolutional,henaff2015deep,levie2018cayleynets,li2018adaptive}, spatial GCNs \cite{hamilton2017inductive,monti2017geometric,gao2018large,niepert2016learning,xu2018powerful,velickovic2018graph}, and recursive graph and tree models  \cite{scarselli2009, diss}. Similar to \cite{Ashraf_Strotherm_Hermes_Hammer_2024}, our proposed architecture employs custom message-passing GNNs that fall under the umbrella of spatial GCNs. 

The task of hydraulic state estimation in WDS involves estimating the pressure head at every node and water flow through every link given the demands at every consumer node and the pressure heads at the reservoirs. EPANET is the prevalent hydraulic simulator that solves a system of equations to solve this task \cite{rossman2020epanet}. A faster alternative method using edge diffusion was recently proposed by \cite{KERIMOV2025122980} with promising results. However, it is unclear how well the method generalizes to changes in input features. A few DL surrogate models for hydraulic state estimation already exist. Reference \cite{xing2022stateestimation} combines the inputs required by EPANET with sparse sensor measurements to solve this task. However, the efficacy of their method is indeterminate given their limited empirical evaluation. Using the same input data as \cite{xing2022stateestimation}, \cite{KERIMOV2024121933} proposed surrogate metamodels using edge-based GNNs. Although they demonstrate good results, their approach is limited to only single-reservoir WDSs. Reference \cite{Ashraf_Strotherm_Hermes_Hammer_2024} are the first to solve the task of state estimation in WDS using a DL model showing promising results. However, their approach suffers from limited generalizability to unseen input features (demands, pipe diameters, etc.). Moreover, since they only experimented with rather small WDSs, it is unclear how well their model scales to larger and more realistic WDSs. 

WDSs are characterized by different features (pressures, demands, flows, pipe attributes, etc.) having values of different magnitudes that are tied together through physical laws. Other domains in critical infrastructure like electricity grids also exhibit similar characteristics \cite{ghamizi2024}. Hence, traditional data normalization methods used in DL cannot be directly applied without violating the physical relationships. This constitutes an open problem that is addressed in this paper.

\section{Methodology} 

We propose a DL surrogate model comprising an improved GNN architecture, an adapted physics-informed algorithm incorporating pumps and pressure-reducing valves (PRVs), an innovative training scheme, and a physics-preserving data normalization method for hydraulic state estimation in WDS. Our model performs equally well on both small and larger realistic WDSs and is robust to both unseen in-distribution and out-of-distribution input features. The WDS is modeled as a graph with nodes $V = V_c \cup V_r = \{v_1,\dots,v_{N_n}\}$, comprising junctions (consumers) $V_c$ and reservoirs (lakes, rivers, etc.) $V_r$, and edges $E = E_{pipe} \cup E_{pump} \cup E_{prv} = \{ e_{vu} \; | \; \forall \, v \in V; u \in \mathcal{N}(v) \} = \{e_1,\dots,e_{N_e}\}$, comprising pipes $E_{pipe}$, pumps $E_{pump}$, and PRVs $E_{prv}$. 

\textit{\textbf{Water Hydraulics}}: Before explaining our methodology, we briefly state the \textit{hydraulic principles} governing the state of a WDS. The \textit{true} demands $\mathbf{d}^* = (d_v^*)_{v \in V}$, \textit{true} pressure heads $\mathbf{h}^* = (h_v^*)_{v \in V}$ and \textit{true} water flows $\mathbf{q}^* = (q_e^*)_{e \in E}$ obey the following hydraulic principles:

\begin{enumerate}
    \item The flow direction property dictates that for two neighboring nodes $v,u \in V$, the flow $q_{e_{vu}}$ from node $v$ to $u$ is equal to the negative of the flow $q_{e_{uv}}$ from $u$ to $v$:
    \begin{align}
    \label{align_flowdirection}
        q_{e_{vu}}^* 
        := 
        -q_{e_{uv}}^*.
    \end{align}
    \item The law of conservation of mass postulates that the sum of all flows at a node is equal to the negative of the demand at that node i.e.
    \begin{align}
    \label{align_MassBalance}
        \sum_{u \in \mathcal{N}(v)} q_{e_{vu}}^* = - d_v^*.
    \end{align}
    \item The law of conservation of energy states that the loss of pressure head between two neighboring nodes is related to the flow through the pipe between these nodes as:
    \begin{align}
    \label{align_HeadLoss}
       h_v^* - h_u^* = r_{e_{vu}} \sgn(q_{e_{vu}}^*) |q_{e_{vu}}^*|^x,
    \end{align}
    \begin{align}
    \label{align_ConstantR}
        r_{e_{vu}} = 
        10.667 ~ l_{e_{vu}} 
        \psi_{e_{vu}}^{-4.871} c_{e_{vu}}^{-1.852}
        > 0,
    \end{align}
    \noindent where $x = 1.852$ and $l_{e_{vu}}$, $\psi_{e_{vu}}$ and $c_{e_{vu}}$ are length, diameter and roughness coefficient of the pipe  \cite{rossman2020epanet}.
\end{enumerate}
\begin{figure*}[h!]
\centerline{\includegraphics[width=.9\textwidth, keepaspectratio]{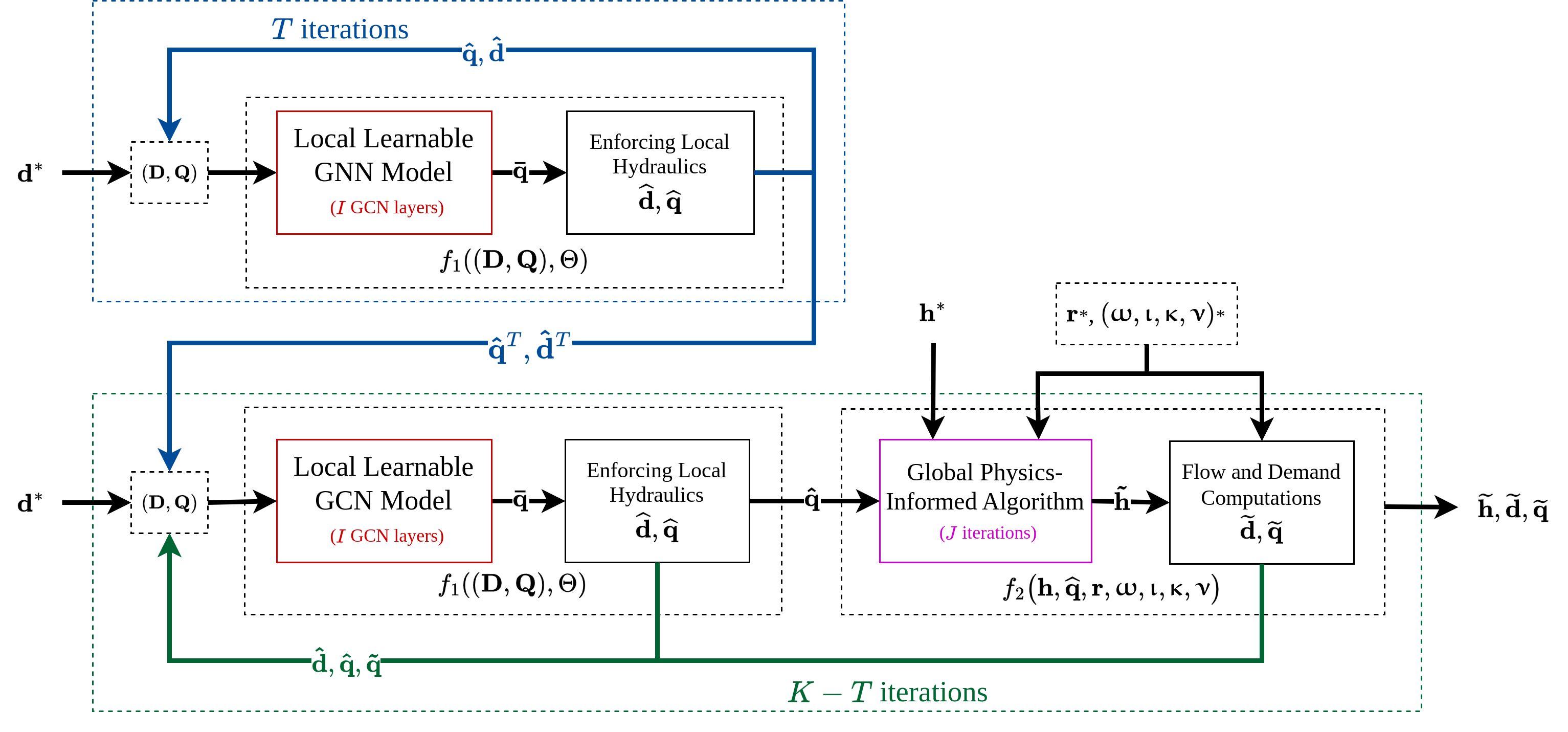}}
\caption{Architecture of the model incorporating a two-phase training scheme.}
\label{fig_architecture}
\end{figure*}
In this work, we integrate two additional types of links into our model -- \textbf{\textit{pumps}} and \textbf{\textit{pressure-reducing valves}} (PRVs). Pumps provide an increase in pressure given the flow as:
\begin{align}
\label{align_PumpHeadGain}
\begin{split}
    h^*_v
    & = 
    h^*_u
    -
    (-\omega^{2} \left(\iota - \kappa \left(-q^*_{e_{vu}}  / \omega)^{\nu} \right) \right),
\end{split}
\end{align}
\noindent where $\iota$ is the pump shutoff head, $\omega$ is the relative speed of the pump, and $\kappa$ and $\nu$ are pump curve coefficients \cite{rossman2020epanet}.
PRVs are used to limit the pressures to specified values $h_{v_{prv_o}}^*$. These head values are applied to the receiving nodes of the PRVs i.e. $V_{prv_o}$, while $V_{prv_i}$ is the set of starting nodes for PRVs.      

\textit{\textbf{Problem Definition}}: The task of hydraulic state estimation in WDS involves estimating the pressure head at every node $(h_v)_{v \in V}$ and water flow through every link $(q_e)_{e \in E}$ given \textit{true} demands at every consumer node $(d_v^*)_{v \in V_c}$, \textit{true} pressure heads at the reservoirs $(h_v^*)_{v \in V_r}$, PRVs pressure head settings $(h_v^*)_{v \in V_{prv_o}}$, pipe attributes $(r_e)_{e \in E_{pipe}}$, and pump attributes $(\omega_e, \iota_e, \kappa_e, \nu_e)_{e \in E_{pump}}$. 

\subsection{Model Architecture}

Our proposed architecture is based on \cite{Ashraf_Strotherm_Hermes_Hammer_2024}. 
GNNs are local operators, while the task of hydraulic state estimation is a global problem. Hence, similar to \cite{Ashraf_Strotherm_Hermes_Hammer_2024}, we combine a local learnable GNN model $f_1$ with a global physics-informed algorithm $f_2$ using an iterative training scheme to solve this task. The model architecture is depicted in Fig. \ref{fig_architecture}.

Node features $\mathbf{D} \in \R^{N_n \times M_n}$ and edge features $\mathbf{Q} \in \R^{N_e \times M_e}$ are inputs to the first component of our model i.e. $f_1$ that estimates flows $\mathbf{\hat{q}} = (\hat{q}_e)_{e \in E} \in \R^{N_e}$ and computes demands $\mathbf{\hat{d}} = (\hat{d}_v)_{v \in V} \in \R^{N_n}$ with $M_n = 3$ and  $M_e = 2$:
\begin{align*}
    f_1((\mathbf{D},\mathbf{Q}), \Theta): &\longmapsto \mathbf{\hat{q}}, \mathbf{\hat{d}},
    \\
    (\mathbf{D},\mathbf{Q}) = ((\mathbf{d}_1, \mathbf{d}_2, \mathbf{d}_3), (\mathbf{q}_1, \mathbf{q}_2)) & = ((\mathbf{d}_v^T)_{v \in V}, (\mathbf{q}_e^T)_{e \in E}).
\end{align*}
\noindent We will explain these features in detail later. The second component involves a physics-informed algorithm similar to \cite{Ashraf_Strotherm_Hermes_Hammer_2024} augmented with adaptations for pumps and PRVs. It takes flows computed by $f_1$ along with \textit{true} pressure heads at the reservoirs and receiving nodes of PRVs $\mathbf{h}^*$, the pipe resistance coefficients $\mathbf{r}^*$, and pump attributes $\boldsymbol{\upomega}^*, \boldsymbol{\upiota}^*, \boldsymbol{\upkappa}^*, \boldsymbol{\upnu}^*$, and computes another set of flows $\mathbf{\tilde{q}}$ and demands $\mathbf{\tilde{d}}$:   
\begin{align*}
\begin{split}
    f_2(\mathbf{h}, \mathbf{\hat{q}}, \mathbf{r}, \boldsymbol{\upomega}, \boldsymbol{\upiota}, \boldsymbol{\upkappa}, \boldsymbol{\upnu}  ) & \longmapsto (\mathbf{\tilde{h}}, \mathbf{\tilde{d}}, \mathbf{\tilde{q}}).
\end{split}    
\end{align*}
Next, we separately describe both components $f_1$ and $f_2$ followed by our training scheme and normalization method.

\subsubsection{Local Learnable GNN-Model}
\label{subsection_LearnableGNNModel}

The first component $f_1$ involves a local learnable GNN model comprising a few GNN layers. Node and edge features are embedded to dimension $M_l$ using linear layers $\alpha$ and $\beta$:
\begin{align*}
    \mathbf{g}_v := \alpha(\mathbf{d}_v), \quad \forall \; v \in V
    \quad  \text{ and }  \quad
    \mathbf{z}_e := \beta(\mathbf{q}_e), \quad \forall \; e \in E.
\end{align*}
Compared to \cite{Ashraf_Strotherm_Hermes_Hammer_2024}, we replace the SeLU activation function with ReLU in the message construction phase of the GNN layers for better convergence. We use $I$ GNN layers, starting with $\mathbf{g}_v^{(0)} = \mathbf{g}_v$ and $\mathbf{z}_e^{(0)} = \mathbf{z}_e$ for all $v \in V$ and $e \in E$, respectively.
For the $i$-th layer for $ i = 0, ..., I-1$, and $e = e_{vu} \in E$, we \textbf{generate messages} by concatenating node and edge embeddings and passing these through a multi-layer perceptron (MLP) $\gamma$:
\begin{align*}
    \mathbf{m}_e^{(i)}
    := 
    \mathrm{ReLU}(\gamma^{(i)}(\mathbf{g}_u^{(i)} \parallel \mathbf{g}_v^{(i)} \parallel \mathbf{z}_e^{(i)}))  
    \text{ with } 
    u \in \mathcal{N}(v).
\end{align*}
\noindent We use max aggregation for \textbf{message aggregation}:
\begin{align*}
    \mathbf{m}_v^{(i)} := \max_{u \in \mathcal{N}(v)} \mathbf{m}_{e_{vu}}^{(i)}, \quad \forall \; v \in V.
\end{align*}

\noindent Node features are \textbf{updated} using an MLP $\eta$:
\begin{align*}
    \mathbf{g}_v^{(i+1)} := \eta^{(i)}(\mathbf{m}_v^{(i)}), \; \forall \; v \in V 
    \text{ and } 
    \mathbf{z}_e^{(i+1)} := \mathbf{m}_e^{(i)}, \; \forall \; e \in E.
\end{align*}

\noindent After the last GNN layer, we concatenate node and edge features and pass these through an MLP $\lambda$:
\begin{align*}
    \mathbf{\bar{z}}_e 
    := 
    \lambda(\mathbf{g}_u^{(I)} \parallel \mathbf{g}_v^{(I)} \parallel \mathbf{z}_e^{(I)})
    \text{ with } 
    u \in \mathcal{N}(v), \; \forall \; e \in E.
\end{align*}

Instead of discarding half of the bidirectional flow embeddings as done by \cite{Ashraf_Strotherm_Hermes_Hammer_2024}, we concatenate the flow embeddings in one direction $\mathbf{\bar{z}}_{e_{in}}$ with the flow embeddings in the other direction $\mathbf{\bar{z}}_{e_{out}}$ and pass these through an MLP $\phi$, to estimate only directed flows $\mathbf{\bar{q}}$ (using residual connections):
\begin{align*}
    \bar{q}_{e_{in}} 
    := 
    \bar{q}_{e_{in}} 
    +
    \phi(\mathbf{\bar{z}}_{e_{in}} \parallel \mathbf{\bar{z}}_{e_{out}})
    , \quad \forall \; e \in E.
\end{align*}

\noindent This allows for full utilization of the model embeddings and faster convergence. 

\textit{Enforcing Local Hydraulics}: We compute flows in the other direction as the negative of the estimated flows using \eqref{align_flowdirection}:
\begin{align*}
    \hat{q}_{e} 
    := 
    \bar{q}_{e_{in}} \parallel (-\bar{q}_{e_{in}}), \quad \forall \; e \in E.
\end{align*}

\noindent The updated demands $\mathbf{\hat{d}} \in \R^{N_n}$ are then computed from the estimated flows $\mathbf{\hat{q}}$ using \eqref{align_MassBalance}.

\subsubsection{Global Physics-Informed Algorithm}
\label{subsection_GlobalPhysicsInformedAlgorithm}

The second component $f_2$ includes a physics-informed message passing algorithm based on the one proposed by \cite{Ashraf_Strotherm_Hermes_Hammer_2024}. We modify the algorithm to adapt to pumps and PRVs. Starting with $\tilde{h}_v^{(0)} = h_v$ and for all $ v \in V \setminus V_r$, we \textit{construct messages} for \textit{\textbf{pipes}} as: 
\begin{align} 
\label{align_EdgeMessageGeneration_PhysicsInformed}
    {m}_e^{(j)}
    := 
    {\tilde{h}}_u^{(j)}
    -
    r_e \left(\mathrm{ReLU}\left(-\hat{q}_e \right) \right)^x, \quad \forall \; e \in E_{pipe},
\end{align}
where $x = 1.852$. For \textit{\textbf{pumps}}, we use \eqref{align_PumpHeadGain} i.e. for all $ e \in E_{pump}$
\begin{align*}
\begin{split}
    {m}_e^{(j)}
    & := 
    {\tilde{h}}_u^{(j)}
    -
    (-\omega_e^{2} \left(\iota_e - \kappa_e (\left(\mathrm{ReLU}\left(-\hat{q}_e \right) \right) / \omega_e)^{\nu} \right).
\end{split}
\end{align*}
Messages are \textit{aggregated} and \textit{updated} by taking the maximum:
\begin{align*}
    {m}_v^{(j)} 
    := 
    \max_{u \in \N(v)} {m}_{e_{vu}}^{(j)}, \quad \forall \; v \in V \setminus V_r,
    \\
    {\tilde{h}}_v^{(j+1)} 
    := 
    \max\{ {\tilde{h}}_v^{(j)}, {m}_v^{(j)} \}, \quad \forall \; v \in V \setminus V_r.
\end{align*}
\noindent where $J$ is the number of iterations that varies based on how close $\mathbf{\hat{q}}$ are to $\mathbf{q^*}$. For the receiving nodes of PRV edges, we define and apply a \textbf{\textit{second update}} that limits the pressure head at these nodes to the values specified by the PRVs:
\begin{align*}
    {\tilde{h}}_v^{(j+1)} 
    := 
    \min\{ {\tilde{h}}_v^{(j)}, {h}_v^{(j)} \} \quad \forall \quad v \in V_{prv_o}.
\end{align*}

\textit{Flow and Demand Computations}: After convergence, we use the reconstructed heads $\mathbf{\tilde{h}} := \mathbf{\tilde{h}}^{(J)} \in \R^{N_n}$ to compute updated flows $\mathbf{\tilde{q}} \in \R^{N_e} $. For all \textit{\textbf{pipes}} $ e \in E_{pipe}$, we use \eqref{align_HeadLoss}:
\begin{align}
\label{align_ComputeFlows}
\begin{split}
    {\tilde{q}}_e
    &:= 
    \sgn({\tilde{h}_v} - {\tilde{h}}_u) 
    (r_e^{-1} |{\tilde{h}_v} - {\tilde{h}}_u|)^{1/x} + \zeta, 
\end{split}
\end{align}
\noindent where $\zeta$ is a very small number added to keep the equation differentiable. For all \textit{\textbf{pumps}} $ e \in E_{pump}$, we use \eqref{align_PumpHeadGain} as:
\begin{align*}
\begin{split}
    {\tilde{q}}_e
    &:= 
    \sgn({\tilde{h}_v} - {\tilde{h}}_u) 
    \left( \max \left\{ 0, \frac{\omega_e^{2}  \iota_e - |\tilde{h}_v - \tilde{h}_u|}
    {\kappa_e \omega_e^{2-\nu}} \right\} \right)^{1/\nu_e}+ \zeta
\end{split}
\end{align*}
\textit{\textbf{PRVs}} introduce a discontinuity in the conservation of energy principle by limiting the pressure to a specific value. Hence, \eqref{align_HeadLoss} cannot be used to compute the flows through the PRVs. This poses a challenge that we address as follows: First, water outflows in the neighborhood of the PRV receiving nodes are summed up (including the \textit{true} demands at those nodes). Then, using the conservation of mass principle \eqref{align_MassBalance}, the flows through the PRVs, which are the inflows to the PRV receiving nodes, are set to the negative of the aforementioned sum:
\begin{align*}
    \textstyle
    {\tilde{q}}_{e} 
    := - 
    \left(\sum_{u \in \mathcal{N}(v) \setminus V_{prv_i} } {\tilde{q}}_{e_{vu}} + d_v^* \right),  \; \forall \, e \in E_{prv}, v \in V_{prv_o}.
\end{align*}
Finally, \eqref{align_MassBalance} is used to compute demands $\mathbf{\tilde{d}} \in \R^{N_n}$: 
\begin{align}
\label{align_ComputeDemands}
    \textstyle
    {\tilde{d}}_v 
    := - 
    \sum_{u \in \mathcal{N}(v)} {\tilde{q}}_{e_{vu}},
    \quad \quad \forall \quad v \in V.
\end{align}
We can write both components of the model as:
\begin{align}
(\mathbf{\tilde{h}}, \mathbf{\tilde{d}}, \mathbf{\tilde{q}}) = f_2(\mathbf{h}, f_1((\mathbf{D}, \mathbf{Q}), \Theta), \mathbf{r}, \boldsymbol{\upomega}, \boldsymbol{\upiota}, \boldsymbol{\upkappa}, \boldsymbol{\upnu}).
\end{align}

\subsection{Training Scheme} \label{subsection_OverallModelandTraining}

Given the pressure heads at reservoirs and receiving nodes of PRVs, demands at consumer nodes, and link attributes, hydraulic state estimation involves estimating flows through links constrained by the pressure heads at nodes. The hydraulic principles explained before dictate the relationships between demands and flows and pressure heads. The idea behind our proposed training scheme is to divide this problem into two phases. In the first phase, GNNs that are local operators attempt to formulate a prior estimate of flows that is not constrained by the law of conservation of energy (cf. \eqref{align_HeadLoss}) and \eqref{align_PumpHeadGain}. Moreover, by applying a few GNN layers multiple times, we are able to condition this prior to the complete neighborhood of each node. In the second phase, we run the global physics-informed algorithm along with the GNN layers to constrain the aforementioned prior to a physically correct state obeying all hydraulic principles. 

Based on this idea, we divide the iterative training scheme into two phases. In the \textbf{first phase}, only $f_1$ is applied for $T \ge \lceil \frac{\Sigma}{I} -1 \rceil$ iterations, where $\Sigma$ is the diameter of the WDS graph. This allows the GNN layers to traverse the complete neighborhood for every node. The only input feature utilized in this phase are the \textit{true} demands $\mathbf{d}^* = (d_v^*)_{v \in V \setminus V_r}$. Hence, we initialize the node features $(\mathbf{d}_1, \mathbf{d}_2, \mathbf{d}_3)$ as:
\begin{align*}
    d_{v1}^{(0)} &:=
    \begin{cases}
    0 \quad & \text{ if } v \in V_r \\
    d_v^* \quad & \text{ if } v \in V \setminus V_r 
    \end{cases}
    ,\\
    d_{v2}^{(0)} &:= \hat{d}_{v}^{(0)} :=
    0, \quad \; \forall \; v \in V 
    ,\\
    d_{v3}^{(0)} &:=
    \begin{cases}
    1 \quad & \text{ if } v \in V_r \\
    0 \quad & \text{ if } v \in V \setminus V_r 
    \end{cases},
\end{align*}
\noindent where $\mathbf{d}_3$ is a reservoir mask that helps the GNN differentiate reservoir nodes from other nodes. We initialize all flows to zero i.e. $\mathbf{\hat{q}} =  \mathbf{\tilde{q}} = 0$. Thus, the initial edge features are:
\begin{align*}
    q_{e1}^{(0)} :=
    \hat{q}_e^{(0)} := 0
    ,
    \qquad q_{e2}^{(0)} :=
    \tilde{q}_e^{(0)} := 0,
    \qquad \forall \; e = e_{vu} \in E.
\end{align*}
\noindent Hence, for the first phase of $T$ iterations, we can write:
{\footnotesize
\begin{align*}
    \mathbf{\hat{q}}^{(t)} &:= f_1\left( \Big((\mathbf{d_1}^{(t)}, \mathbf{d_2}^{(t)}, \mathbf{d_3}^{(t)}),
    (\mathbf{q_1}^{(t)}, \mathbf{q_2}^{(t)})\Big), \Theta \right),\\
    \mathbf{d_1}^{(t+1)} &:= \mathbf{d}^{(0)} \text{ (true demands)},\\
    \mathbf{d_2}^{(t+1)} &:= \mathbf{\hat{d}}^{(t)}\text{ (demands from $f_1$)},\\
    \mathbf{d_3}^{(t+1)} &:= \mathbf{d_3}^{(0)}\text{ (fixed)},\\
    \mathbf{q_1}^{(t+1)} &:= \mathbf{\hat{q}}^{(t)} \text{ (flows from $f_1$)},\\
    \mathbf{q_2}^{(t+1)} &:= \mathbf{\tilde{q}}^{(t)} \text{ (all zeroes)}.
\end{align*}
}

For the \textbf{second phase}, we also utilize the other input features i.e. pressure heads at the reservoirs and receiving nodes of PRVs $\mathbf{h}$, the pipe resistance coefficients $\mathbf{r}$, and pump attributes $\boldsymbol{\upomega}, \boldsymbol{\upiota}, \boldsymbol{\upkappa}, \boldsymbol{\upnu}$, which are initialized as:
\begin{align*}
    h_v^{(0)} &:=
    \begin{cases}
     h_v^* \quad & \text{ if } v \in V_r \\
     h_{v_{prv_o}}^* \quad & \text{ if } v \in V_{prv_o} \\
    0 \quad & \text{ if } v \in V \setminus (V_r \cup V_{prv_o}) 
    \end{cases}
    ,\\
    r_e^{(0)} & :=
     \quad r_e^*, \qquad \qquad \qquad \;\, \forall \quad e \in E_{pipe} 
     ,\\
    (\omega_e, \iota_e, \kappa_e, \nu_e)^{(0)} & :=
     \quad (\omega_e, \iota_e, \kappa_e, \nu_e)^*, \quad \forall \quad e \in E_{pump}.
\end{align*}

\noindent Here, we run both $f_1$ and $f_2$ for another $P = K - T$ iterations, where $K$ is the total number of iterations that is a hyperparameter. For this phase, we can write the update as:

{\footnotesize
\begin{align*}
    \mathbf{\hat{q}}^{(p)} &:= f_1\left( \Big((\mathbf{d_1}^{(p)}, \mathbf{d_2}^{(p)}, \mathbf{d_3}^{(p)}),
    (\mathbf{q_1}^{(p)}, \mathbf{q_2}^{(p)})\Big), \Theta  \right),\\
    (\mathbf{\tilde{h}}^{(p)}, \mathbf{\tilde{d}}^{(p)}, \mathbf{\tilde{q}}^{(p)}) &:= f_2(\mathbf{h}^{(0)},\mathbf{\hat{q}}^{(p)}, \mathbf{r}^{(0)}, (\boldsymbol{\upomega}, \boldsymbol{\upiota}, \boldsymbol{\upkappa}, \boldsymbol{\upnu})^{(0)}),\\
    \mathbf{d_1}^{(p+1)} &:= \mathbf{d}^{(0)} \text{ (true demands)},\\
    \mathbf{d_2}^{(p+1)} &:= \mathbf{\hat{d}}^{(p)}\text{ (demands from $f_1$)},\\
    \mathbf{d_3}^{(p+1)} &:= \mathbf{d_3}^{(0)}\text{ (fixed)},\\
    \mathbf{q_1}^{(p+1)} &:= \mathbf{\hat{q}}^{(p)} \text{ (flows from $f_1$)},\\
    \mathbf{q_2}^{(p+1)} &:= \mathbf{\tilde{q}}^{(p)} \text{ (flows from $f_2$)},\\
    \mathbf{h}^{(p+1)} &:= \mathbf{h}^{(0)}\text{ (fixed)},\\
    \mathbf{r}^{(p+1)} &:= \mathbf{r}^{(0)}\text{ (fixed)}, \\
    (\boldsymbol{\upomega}, \boldsymbol{\upiota}, \boldsymbol{\upkappa}, \boldsymbol{\upnu})^{(p+1)} &:= (\boldsymbol{\upomega}, \boldsymbol{\upiota}, \boldsymbol{\upkappa}, \boldsymbol{\upnu})^{(0)}\text{ (fixed)}.
\end{align*}
}
\subsection{Objective Function}

After completing both the aforementioned two-phase forward pass, we back-propagate to update the model parameters using the following loss function with three terms \cite{Ashraf_Strotherm_Hermes_Hammer_2024}:
\begin{align}
\label{align_loss}
    \mathcal{L} = 
    \mathcal{L} (\mathbf{d}^*, \mathbf{\hat{d}}^{(K)}) 
    + \rho
    \mathcal{L} (\mathbf{d}^*, \mathbf{\tilde{d}}^{(K)}) 
    + \delta
    \mathcal{L} (\mathbf{\hat{q}}^{(K)}, \mathbf{\tilde{q}}^{(K)}),
\end{align}
\noindent where $\mathcal{L}$ denotes the L1 loss and $\rho$ and $\delta$ are hyperparameters.

\subsection{Data Normalization}
\label{sec_data_normalization}

Water pressure heads, flows, demands, and pipe attributes have values of different magnitudes that are tied together through hydraulic principles. Since demands and pipe attributes are the major inputs to the model, normalizing these with traditional data normalization methods used in DL will violate the hydraulic principles. We address this problem by proposing a unique physics-preserving data normalization method where we:

\begin{enumerate}
    \item normalize both demands and pipe attributes within specific ranges eradicating exploding gradients and allowing for generalization to out-of-distribution values,
    \item adjust the reservoir heads, PRV settings, and pump attributes using the hydraulic principles ensuring adherence to physics and making this normalization reversible.     
\end{enumerate}


To normalize demands, we divide these by their sum i.e.
\begin{align*}
    \textstyle
    d_{sum}^{'*} :=
    \sum_{v \in V \setminus V_r} d_{v}^{'*}, 
    \quad  
    d_{v}^{*} :=
    {d_{v}^{'*}} / {d_{sum}^{'*}}, 
    \quad \forall \; v \in V \setminus V_r,
\end{align*}
\noindent where $'$ denotes non-normalized values. This keeps the demand values below $1$ but at the same time limits the largest flow to around $1$. This is because the largest flow in a WDS is close to the sum of all demands in the WDS. Since $f_1$ estimates flows $\mathbf{\hat{q}}$, which in turn are used by $f_2$ to compute flows $\mathbf{\tilde{q}}$, keeping these values between $0$ and around $1$ eradicates exploding gradients to a great extent (cf. Sec. \ref{EradicatingExplodingGradients}).

Adding noise to the diameters sampled from a normal distribution does not guarantee that the induced variations in the model input feature $\mathbf{r}$ follow the same distribution. In order to undermine the outliers, we normalize $\mathbf{r}$ by dividing it by three times its standard deviation across the WDS i.e. 
\begin{align*}
    r_{fac}^{'*} :=
    (3 * \sigma_{r_{e}^{'*}}) / {\tau},  
    \quad \quad 
    r_{e}^{*} :=
    {r_{e}^{'*}} / {r_{fac}^{'*}},  \quad \forall \;  e \in E_{pipe},
\end{align*}
\noindent where $\tau=1000$ is a scaling factor. To preserve the hydraulic relationships dictated by \eqref{align_HeadLoss} and \eqref{align_PumpHeadGain}, we must also change the reservoir heads, PRV settings, and pump attributes as follows:
\begin{align*}
    h_{v}^{*} &:=
    {h_{v}^{'*}} / ((d_{sum}^{'*})^x  r_{fac}^{'*}), \qquad \forall \; v \in V_r \cup V_{{prv}_o},
    \\
    \iota_{e}^{*} &:=
    \iota_{e}^{'*} / (r_{fac}^{'*} \, (d_{sum}^{'*})^x), \qquad \forall \; e \in E_{pump},
    \\
    \kappa_{e}^{*} &:=
    (\kappa_{e}^{'*} \, (d_{sum}^{'*})^{\nu - x})  / r_{fac}^{'*} , \;\;\,  \forall \; e \in E_{pump}.
\end{align*}

Our normalization method ensures that the estimated state of the WDS by the model ($\mathbf{\tilde{h}}, \mathbf{\tilde{q}}, \mathbf{\tilde{d}}$) obeys all hydraulic principles and is completely reversible. To get back to the de-normalized state ($\mathbf{\tilde{h}}^{'}, \mathbf{\tilde{q}}^{'}, \mathbf{\tilde{d}}^{'}$), we de-normalize demands as:
\begin{align*}
    \tilde{d}_{v}^{'} :=
    \tilde{d}_{v} \, d_{sum}^{'*}, \quad \forall \; v \in V \setminus V_r.
\end{align*}
Using the linear relationship between flows and demands (cf. \eqref{align_MassBalance}), we de-normalize flows as follows:
\begin{align*}
    \tilde{q}_{e}^{'} :=
    \tilde{q}_{e} \, d_{sum}^{'}, \quad \forall \; e \in E.
\end{align*}
The relationship between heads and flows is nonlinear. Hence, we de-normalize heads leveraging \eqref{align_HeadLoss} as follows:
\begin{align*}
    \tilde{h}_{v}^{'} :=
    \tilde{h}_{v} * (d_{sum}^{'*})^x * r_{fac}^{'*}, \quad \forall \; v \in V. 
\end{align*}

\section{Experiments}

We evaluate our model on eleven WDSs. First, we explain the datasets and the training setup. Next, we demonstrate the accuracy and scalability of our approach to larger WDSs in comparison with the hydraulic simulator and the SOTA model from \cite{Ashraf_Strotherm_Hermes_Hammer_2024}. Then, we analyze the robustness of our model to changes in input features followed by the efficiency gains obtained using our approach.

\begin{table*}[!t]
\caption{Properties/Characteristics of WDSs}
\begin{center}
\begin{tabular}{|c|c|c|c|c|c|c|c|}
\hline
 \textbf{Attributes} & \textbf{No. of junctions} & \textbf{No. of reservoirs} & \textbf{No. of pipes} & \textbf{No. of pumps} & \textbf{No. of PRVs} & \textbf{Diameter} & \textbf{Node degree}  \\
\textbf{WDS} &  &  &  &  &  &  & \textbf{(min, mean, max)}  \\
\hline
\textbf{Anytown} & 19 & 3 & 40 & 1 & 0 & 5 & (2, 7.45, 14)  \\
\hline
\textbf{Hanoi} & 31 & 1 & 34 & 0 & 0 & 13 & (2, 4.25, 8)  \\
\hline
\textbf{Pescara} & 68 & 3 & 99 & 0 & 0 & 20 & (2, 5.52, 10) \\
\hline
\textbf{L-Town Area C} & 92 & 1 & 109 & 0 & 0 & 20 & (2, 4.69, 8) \\
\hline
\textbf{Zhi Jiang} & 113 & 1 & 164 & 0 & 0 & 24 & (2, 5.75, 8) \\
\hline
\textbf{Modena} & 268 & 4 & 317 & 0 & 0 & 38 & (2, 4.66, 10) \\
\hline
\textbf{PA1} & 337 & 2 & 399 & 0 & 0 & 55 & (2, 4.71, 10) \\
\hline
\textbf{Balerma} & 443 & 4 & 454 & 0 & 0 & 60 & (2, 4.06, 10) \\
\hline
\textbf{L-Town Area A} & 659 & 2 & 764 & 0 & 2 & 79 & (2, 4.64, 10)  \\
\hline
\textbf{L-Town All Areas} & 782 & 3 & 905 & 1 & 3 & 79 & (2, 5.42, 10)  \\
\hline
\textbf{KL} & 935 & 1 & 1274 & 0 & 0 & 53 & (2, 5.42, 10)  \\
\hline

\end{tabular}
\label{wdss_attributes}
\end{center}
\end{table*}

\begin{table}[!b]
\caption{Hyperparameters}
\begin{center}
\begin{tabular}{|c|c|c|c|}
\hline
& \textbf{Phase 1 iterations T} & \multicolumn{2}{|c|}{\textbf{Range of K}}  \\
\hline
\textbf{WDS} & \textbf{SPI-GNN} & \textbf{PI-GNN} & \textbf{SPI-GNN}  \\
\hline
\textbf{Anytown} & 0  & N/A & [5, 10]      \\
\hline
\textbf{Hanoi} & 2 & [7, 12] & [10, 15]      \\
\hline
\textbf{Pescara} & 3 & [8, 13] & [10, 15]    \\
\hline
\textbf{L-Town Area C} & 3 & [8, 13] & [10, 15]     \\
\hline
\textbf{Zhi Jiang} & 4 & [9, 14] & [10, 15]     \\
\hline
\textbf{Modena} & 7 & [12, 17] & [12, 17]      \\
\hline
\textbf{PA1} & 10 & [15, 20] & [15, 20]    \\
\hline
\textbf{Balerma} & 11 & [16, 21] & [16, 21]     \\
\hline
\textbf{L-Town Area A} & 15 & N/A & [20, 25]     \\
\hline
\textbf{L-Town All Areas} & 15 & N/A & [20, 25]     \\
\hline
\textbf{KL} & 10 & N/A & [15, 20]     \\
\hline

\end{tabular}
\label{hyperparameters}
\end{center}
\end{table}

\begin{table*}[!t]
\caption{Mean absolute error (MAE) on test datasets. The best values are highlighted in bold.}
\begin{center}
\begin{tabular}{|c|c|c|c|c|c|c|}
\hline
\textbf{MAE }(x $10^{-2}$) & 
\multicolumn{2}{|c|}{\textbf{Demands (Estimated vs True) }} & 
\multicolumn{2}{|c|}{\textbf{Flows (Estimated vs Simulator) }} &
\multicolumn{2}{|c|}{\textbf{Heads (Estimated vs Simulator) }} \\
\hline
\textbf{WDSs} & \textbf{PI-GNN} & \textbf{SPI-GNN} & \textbf{PI-GNN} & \textbf{SPI-GNN} & \textbf{PI-GNN} & \textbf{SPI-GNN} \\
\hline

\textbf{Anytown} & N/A & 0.019 $\pm$ 0.029 & N/A & 0.005 $\pm$ 0.009 & N/A & 0.001 $\pm$ 0.001 \\
\hline
\textbf{Hanoi} & 0.018 $\pm$ 0.044 & \textbf{0.012} $\pm$ 0.023 & 0.001 $\pm$ 0.002 & \textbf{0.001} $\pm$ 0.002 & \textbf{0.004} $\pm$ 0.004 & 0.013 $\pm$ 0.005 \\
\hline
\textbf{Pescara} & 0.078 $\pm$ 0.189 & \textbf{0.068} $\pm$ 0.129 & 0.012 $\pm$ 0.022 & \textbf{0.011} $\pm$ 0.017 & \textbf{0.049} $\pm$ 0.065 & 0.052 $\pm$ 0.068 \\
\hline
\textbf{L-Town Area C} & \textbf{0.117} $\pm$ 0.243 & 0.126 $\pm$ 0.025 & \textbf{0.004} $\pm$ 0.008 & 0.005 $\pm$ 0.007 & 0.057 $\pm$ 0.040 & \textbf{0.028} $\pm$ 0.024 \\
\hline
\textbf{Zhi Jiang} & \textbf{0.075} $\pm$ 0.123 & 0.090 $\pm$ 0.160 & 0.004 $\pm$ 0.007 & \textbf{0.003} $\pm$ 0.005 & 0.084 $\pm$ 0.062 & \textbf{0.028} $\pm$ 0.019
\\
\hline
\textbf{Modena} & 1.025 $\pm$ 2.642 & \textbf{0.361} $\pm$ 1.874 & 0.061 $\pm$ 0.119 & \textbf{0.019} $\pm$ 0.085 & 0.361 $\pm$ 0.301 & \textbf{0.124} $\pm$ 0.121 \\
\hline
\textbf{PA1} & 1.201 $\pm$ 4.109 & \textbf{0.264} $\pm$ 0.773 & 0.297 $\pm$ 0.576 & \textbf{0.043} $\pm$ 0.098 & 2.777 $\pm$ 1.854 & \textbf{0.344} $\pm$ 0.273 \\
\hline
\textbf{Balerma} & 0.224 $\pm$ 0.918 & \textbf{0.092} $\pm$ 0.442 & 0.012 $\pm$ 0.032 & \textbf{0.001} $\pm$ 0.002 & 0.408 $\pm$ 0.419 & \textbf{0.007} $\pm$ 0.006 \\
\hline
\textbf{L-Town Area A} & N/A & 0.761 $\pm$ 1.962 & N/A & 0.071 $\pm$ 0.115 & N/A & 0.125 $\pm$ 0.077 \\
\hline
\textbf{L-Town All Areas} & N/A & 0.278 $\pm$ 0.668 & N/A & 0.028 $\pm$ 0.050 & N/A & 0.024 $\pm$ 0.023 \\
\hline
\textbf{KL} & N/A & 0.566 $\pm$ 2.501 & N/A & 0.033 $\pm$ 0.193 & N/A & 0.534 $\pm$ 0.371 \\
\hline

\end{tabular}
\label{test_eval_results}
\end{center}
\end{table*}

\subsection{Datasets}

WDSs consist of different types of nodes and links. Consumer nodes are called junctions, large water sources like lakes and rivers are reservoirs, and artificial water sources are water tanks. The majority of links in a WDS are water pipes. At some locations, different kinds of valves and pumps are used to maintain water pressure. We use eleven WDSs ranging from small to significantly larger WDSs. The WDS configuration files of these WDSs are obtained from \cite{epanet_collection} and \cite{github:water_benchmark_hub}. Key properties/characteristics of these WDSs are listed in Table \ref{wdss_attributes}.   

The data contained in the configuration files are not enough to train a generalizable DL model. These files normally contain a single set of node demands and pipe attributes ($l$, $\psi$, $c$). In some cases, a few demand patterns are available that are still insufficient for training. Hence, for every WDS, we create datasets by sampling the input features from normal distributions. Robustness to varying demands and pipe attributes is important for a surrogate model. Among pipe attributes, diameters are of the utmost importance given their inverse exponential relationship to the WDS state ($\mathbf{h}, \mathbf{q}, \mathbf{d}$) (cf. \eqref{align_ConstantR}).  

Unlike typical GNNs, we do not need a huge amount of training data because the design of the loss function incorporates hydraulic principles in the training of the model \eqref{align_loss}.
Therefore, for every WDS, we create a total of $6000$ samples. The diameters are varied every 6th sample while demands are varied across all samples. For demands, we create demand patterns by sampling from a normal distribution $\mathcal{N}(1,\,0.1)$ and adding an offset also sampled from the same distribution $\mathcal{N}(1,\,0.1)$. This ensures different patterns for different nodes. These patterns are multiplied by the base demands to get different demands across samples. We add noise to the base diameters by sampling from a normal distribution $\mathcal{N}(0,\,0.01)$. We use a much smaller standard deviation for diameters because the WDS structure is much more sensitive to changes in diameters than to changes in demands (cf. \eqref{align_MassBalance}, \ref{align_HeadLoss}, \ref{align_ConstantR}). 

\subsection{Eradicating Exploding Gradients}
\label{EradicatingExplodingGradients}

Our model does not suffer from the problem of exploding gradients in the same way as \cite{Ashraf_Strotherm_Hermes_Hammer_2024}. Applying $f_2$ repeatedly means applying \eqref{align_EdgeMessageGeneration_PhysicsInformed} and \eqref{align_ComputeFlows} repeatedly. This causes the gradients to explode because of the power term in \eqref{align_EdgeMessageGeneration_PhysicsInformed} and the root term in \eqref{align_ComputeFlows}. 
The maximum flow in a WDS is close to the sum of all demands. Hence, the exploding gradients caused by \eqref{align_EdgeMessageGeneration_PhysicsInformed} are eradicated since we use the sum of demands for normalization restricting the largest flow value to around $1$. Moreover, we split the training into two phases. In the first phase, only $f_1$ is run that does not involve these equations. In the second phase, both $f_1$ and $f_2$ are run within a fixed range of iterations for all WDSs. This limits the extent of exploding gradients caused by \eqref{align_ComputeFlows} unlike \cite{Ashraf_Strotherm_Hermes_Hammer_2024}, where both $f_1$ and $f_2$ were run for all iterations. Nevertheless, the gradients still assume large values due to the iterative training scheme. Hence, we use gradient norm clipping for smoother training.

\subsection{Training Setup}

The datasets are divided into 60:20:20 train-validation-test splits ($3600$ samples for training and $1200$ samples each for validation and testing). All models are implemented in Pytorch. Adam optimizer is used for training and no bias is used in any of the layers. We use only five GNN layers $I$ for all WDSs. All MLPs consist of only one layer. The number of iterations for the first phase $T$ is set equal to $\lceil \frac{\Sigma}{I} -1 \rceil$, while the number of iterations for the second phase $K - T$ is varied between a fixed range as given in Table \ref{hyperparameters}. The latent dimension is of size $128$ and both $\rho$ and $\delta$ are set to $0.1$. All models are trained for $1500$ epochs. A learning rate scheduler is used that starts with a learning rate of $0.0001$ that is reduced every $150$ epochs by a factor of $0.75$. Gradient clipping is used by clipping the norm of the gradients to $0.00001$. Tanks in PA1 and L-Town are converted to reservoirs (cf. Sec. \ref{LimitationsandFutureWork}). For comparison with the hydraulic simulator, WNTR python library is used for running simulations \cite{klise2018overview}.

\subsection{Results and Analysis}

In this section, we present the results of our experiments and evaluate all results using a normalized mean absolute error (MAE). We apply min-max normalization to all state features ($\mathbf{\tilde{h}}^{'}, \mathbf{\tilde{q}}^{'}, \mathbf{\tilde{d}}^{'}$) using the min and max from the corresponding \textit{true} values ($\mathbf{{h}^*}^{'}, \mathbf{{q}^*}^{'}, \mathbf{{d}^*}^{'}$). Then the MAE is computed on these normalized features. This allows us to compare features with different magnitudes using the same metric.

\subsubsection{Evaluation on Test Dataset}

We evaluate trained models for each WDS on the test split comprising $1200$ samples. We compare our SPI-GNN model and the PI-GNN model of \cite{Ashraf_Strotherm_Hermes_Hammer_2024} with \textit{true} demands, and flows and heads from the hydraulic simulator. The results are given in Table \ref{test_eval_results}. For small WDSs, SPI-GNN achieves similar MAE as the PI-GNN despite using fewer number of iterations (Table \ref{hyperparameters}). For larger WDSs, SPI-GNN clearly outperforms PI-GNN. We were unable to complete the training of PI-GNN for some WDSs due to pumps, PRVs, and exploding gradients. Table \ref{test_eval_results} shows the generalizability of our model to unseen data from the same distribution as the training data. However, a surrogate model is vastly more useful when it can also generalize to out-of-distribution data as detailed in the following sections.

\subsubsection{Robustness w.r.t. Demands}

\begin{figure}[!t]
\centerline{\includegraphics[width=\columnwidth, keepaspectratio]{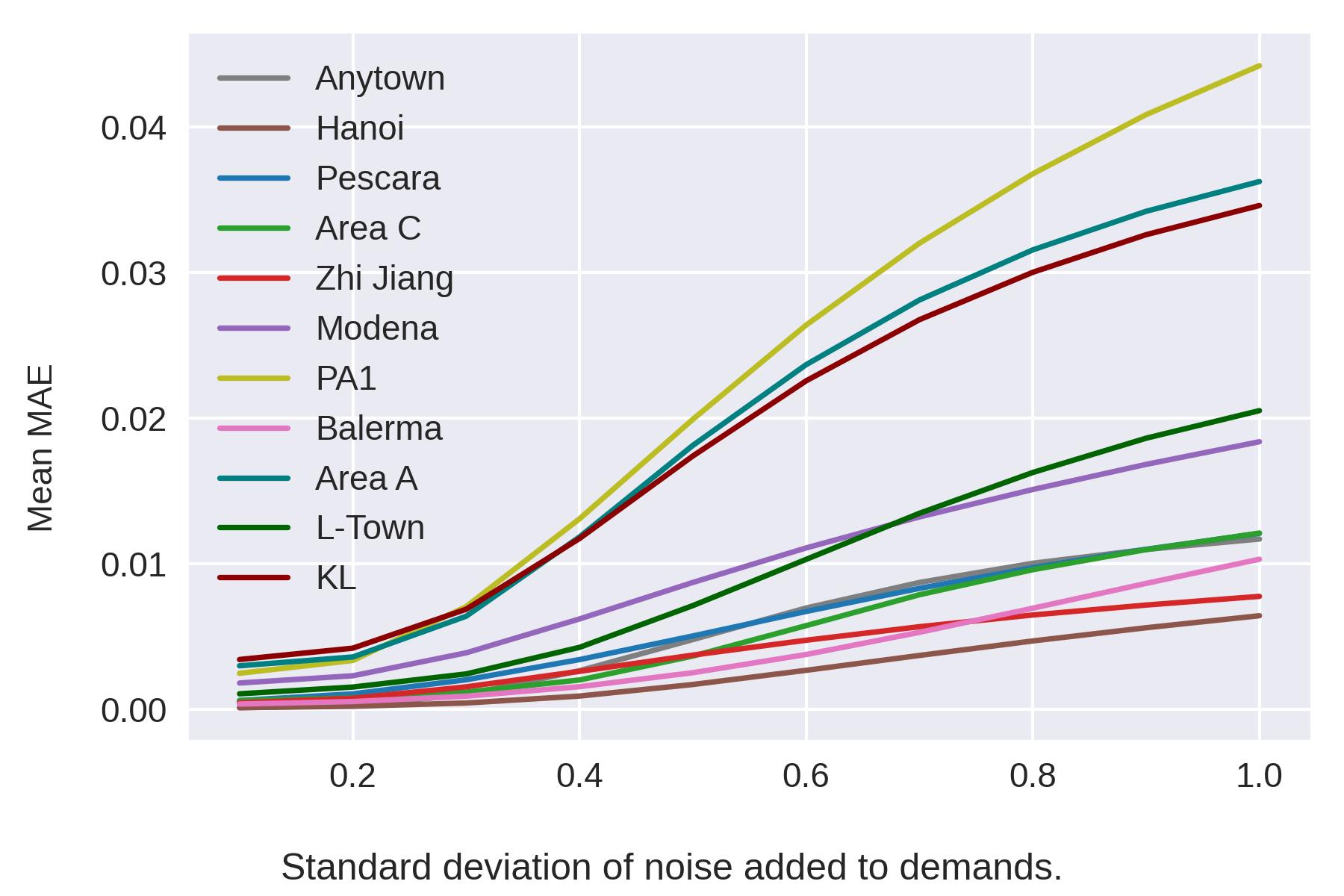}}
\caption{Robustness of SPI-GNN to change in demands.}
\label{fig_demands_robustness}
\end{figure}

The demand patterns in the training dataset are sampled from a normal distribution $\mathcal{N}(1,\,0.1)$. In order to evaluate the robustness of our model, we increase the standard deviation of the aforementioned distribution from $0.1$ to $1.0$. Hence, we evaluate how well our model performs to unseen demands as much as ten times higher than the ones that the model has seen in the training. For this, we generate $1000$ samples for each standard deviation ranging from $0.1$ to $1.0$ with a step size of $0.1$. The process is repeated $10$ times using different random seeds and means are computed. We evaluate MAE on the whole WDS state i.e. for heads, flows, and demands. Since we use a normalized MAE, we average the MAE on all three features to present a combined mean MAE in Fig. \ref{fig_demands_robustness}. As can be seen, the mean MAE stays below $0.05$ even for larger WDSs and the biggest increase in demands. More importantly, the error stays at a similar level as in the training for up to three times increase in the demands. These results demonstrate the robustness of our proposed method to unseen increase in demands.

\begin{figure}[!t]
\centerline{\includegraphics[width=\columnwidth, keepaspectratio]{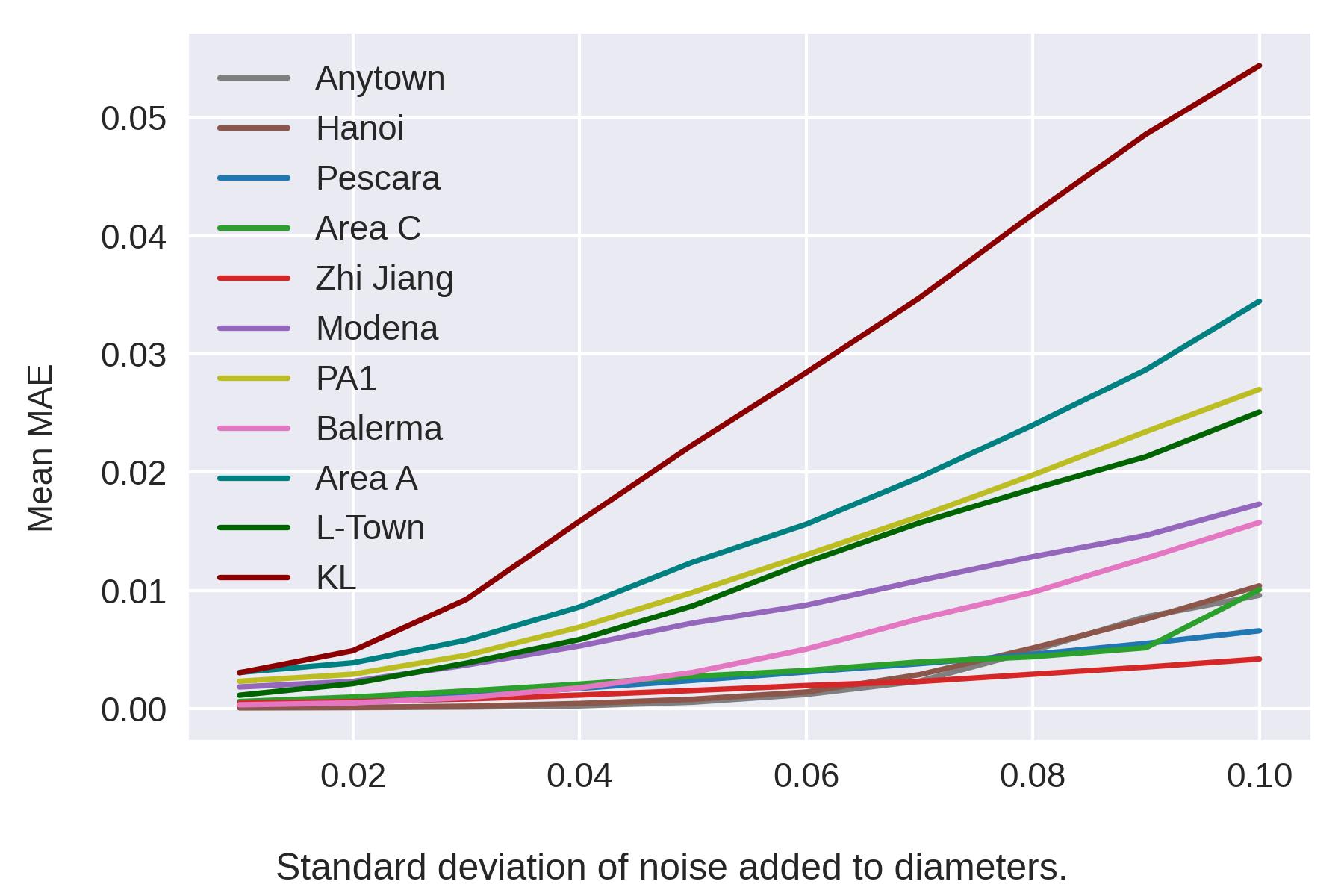}}
\caption{Robustness of SPI-GNN to change in pipe diameters.}
\label{fig_dias_robustness}
\end{figure}

\subsubsection{Robustness w.r.t. Pipe Diameters}

The noise added to the diameters in the training dataset is sampled from a normal distribution $\mathcal{N}(0,\,0.01)$. Similar to the case of demands, in order to get out-of-distribution unseen diameters, we generate 1000 samples for each standard deviation ranging from 0.01 to 0.1 with a step size of 0.01. The process is repeated 10 times using different random seeds and mean MAE is plotted in Fig. \ref{fig_dias_robustness}. The results are very similar to the ones in case of demands exhibiting similar mean MAEs. Hence, our method is also robust to unseen changes in pipe diameters.

\begin{table}[!b]
\caption{Percentage speed-up obtained by SPI-GNN compared to the Hydraulic Simulator in case of increasing demands.}
\begin{center}
\begin{tabular}{|c|c|c|c|c|c|}
\hline
\textbf{Standard} & \textbf{KL} & \textbf{PA1} & \textbf{Balerma} & \textbf{Modena} & \textbf{Pescara}  \\
\textbf{Deviation} & & & & & \\
\hline
\textbf{0.1} & 190 & 557 & 1,221  & 1,640 & 2,651      \\
\hline
\textbf{0.2} & 245 & 1,086 & 2,595 & 3,280 & 6,647      \\
\hline
\textbf{0.3} & 287 & 1,713 & 3,942 & 4,834 & 9,272    \\
\hline
\textbf{0.4} & 339 & 2,218 & 5,465 & 6,299 & 11,328     \\
\hline
\textbf{0.5} & 375 & 2,713 & 6,881 & 7,194 & 13,708     \\
\hline
\textbf{0.6} & 407 & 3,104 & 8,183 & 8,471 & 13,742     \\
\hline
\textbf{0.7} & 437 & 3,432 & 9,379 & 9,219 & 16,595     \\
\hline
\textbf{0.8} & 456 & 3,775 & 10,565 & 10,410 & 18,507     \\
\hline
\textbf{0.9} & 492 & 4,013 & 11,459 & 11,252 & 20,098     \\
\hline
\textbf{1.0} & 523 & 4,286 & 12,641 & 11,619 & 21,335     \\
\hline

\end{tabular}
\label{speedup}
\end{center}
\end{table}

\subsubsection{Efficiency Gains}

The efficiency gains in terms of faster evaluation times of a surrogate DL model were highlighted by \cite{Ashraf_Strotherm_Hermes_Hammer_2024}. Here, we re-emphasize this by showing that a DL model is even more useful when a WDS has multiple reservoirs. Since state estimation in WDS is a complex problem, we observe that the increase in demands makes this problem much more difficult for the simulator as it needs more time to solve it when the WDS has more than one reservoir. On the contrary, our DL model can solve it with the same efficiency as shown in Table \ref{speedup}. As can be seen, with the increase in the standard deviation of the noise added to the demands, the percentage speed-up obtained by SPI-GNN for multi-reservoir WDSs (PA1, Balerma, Modena, Pescara) increases aggressively compared to the single-reservoir WDS (KL).

\section{Limitations and Future Work}
\label{LimitationsandFutureWork}

Currently, all DL methods including our approach solve the problem of hydraulic state estimation as static in time i.e. each sample is solved independently. This is possible as long as there are no tanks in a WDS. In WDSs with tanks, tank levels and the corresponding pressure heads, change with time depending on the water flow in/out of the tank. This makes the problem temporal, which we aim to solve in the future.  

Like any other DL model, our model is completely differentiable. Hence, it can be repurposed to solve other tasks in the WDS domain using gradient methods. Some of these tasks include optimal sensor placement, WDS rehabilitation through pipe features optimization, and estimating pressure heads everywhere in a WDS using only sparse sensors. We plan to evaluate these tasks in future work.

\section{Conclusion}

We propose a new DL surrogate model for hydraulic state estimation in WDSs comprising an improved architecture and an innovative training scheme. Using a physics-preserving data normalization method, we demonstrate that our model scales well to larger WDSs. Moreover, it is highly robust to out-of-distribution input features. Furthermore, with the ability to incorporate more complex physical components of WDSs like pumps and PRVs, our approach constitutes a big step towards reducing the gap between simulation and the real world. Due to immense efficiency gains, our model is ideally suitable for WDS planning, expansion, and rehabilitation, where numerous simulations are required. Overall, we believe that our proposed method is a significant contribution to the use of AI for WDSs.

\bibliographystyle{IEEEtran}
\bibliography{IEEEabrv, references}

\end{document}